\pdfoutput=1
\documentclass[11pt]{article}

\usepackage[final]{acl}

\usepackage{times}
\usepackage{latexsym}

\usepackage[T1]{fontenc}
\usepackage[utf8]{inputenc}

\usepackage{microtype}

\usepackage{inconsolata}
\usepackage{graphicx}
\usepackage{caption}
\usepackage{subcaption}
\usepackage{diagbox}

\usepackage{tabularx}
\usepackage{multirow}
\usepackage{tablefootnote}
\usepackage{algorithm}
\usepackage{algpseudocode}
\usepackage{enumerate}

\usepackage{CJKutf8}
\usepackage{hyperref}

\title{The Model Arena for Cross-lingual Sentiment Analysis: A Comparative Study in the Era of Large Language Models}

\author{Xiliang Zhu, \ Shayna Gardiner, \ Tere Roldán,  \ David Rossouw\\
        Dialpad Inc.\\
        \texttt{\{xzhu, sgardiner, tere.roldan, davidr\}@dialpad.com} \\}

\begin{document}
\maketitle
\begin{abstract}
Sentiment analysis serves as a pivotal component in Natural Language Processing (NLP). Advancements in multilingual pre-trained models such as XLM-R \cite{conneau-etal-2020-unsupervised} and mT5 \cite{xue-etal-2021-mt5} have contributed to the increasing interest in cross-lingual sentiment analysis. The recent emergence in Large Language Models (LLM) has significantly advanced general NLP tasks, however, the capability of such LLMs in cross-lingual sentiment analysis has not been fully studied. This work undertakes an empirical analysis to compare the cross-lingual transfer capability of public Small Multilingual Language Models (SMLM) like XLM-R, against English-centric LLMs such as Llama-3 \cite{llama3modelcard}, in the context of sentiment analysis across English, Spanish, French and Chinese. Our findings reveal that among public models, SMLMs exhibit superior zero-shot cross-lingual performance relative to LLMs. However, in few-shot cross-lingual settings, public LLMs demonstrate an enhanced adaptive potential. In addition, we observe that proprietary GPT-3.5 \footnote{\url{https://platform.openai.com/docs/models/gpt-3-5-turbo}} and GPT-4 \cite{openai2024gpt4} lead in zero-shot cross-lingual capability, but are outpaced by public models in few-shot scenarios.
\end{abstract}

\section{Introduction}
Sentiment analysis has received considerable attention over the years in the field of Natural Language Processing (NLP) due to its profound value in both academic research and industry applications. Traditionally, studies in sentiment analysis had been mostly focused on high-resource languages such as English due to a deficit of annotated data in other low-resource languages, but recent research has emerged to address this issue by leveraging machine translation to augment data resources \cite{ARAUJO20201078} \cite{joshi-etal-2020-state}. 

Besides the research efforts in producing multilingual datasets for sentiment analysis, multilingual model architectures have become increasingly popular since the introduction of multilingual pre-trained language models such as mBERT \cite{devlin-etal-2019-bert}, XLM-R \cite{conneau-etal-2020-unsupervised} and mT5 \cite{xue-etal-2021-mt5} and BLOOM \cite{bigscience_workshop_2022}. Such multilingual pre-trained language models exploit the power of large-scale unsupervised textual data from a mixture of many languages, facilitating zero-shot and few-shot cross-lingual transfer from a source to a target language on different downstream NLP tasks, albeit with varying performance outcomes \cite{lauscher-etal-2020-zero}. 

More recently, Large Language Models (LLM) such as GPT-3 \cite{brown2020language}, Llama-2 \cite{touvron2023llama} and Llama-3 \cite{llama3modelcard} have collected immense attention for their unparalleled performance in text generation. \cite{zhang2023sentiment} shows the strong capability of LLMs with few-shot in-context learning in public English sentiment analysis tasks. Although most of the LLMs are pre-trained using corpora with a dominant presence of English, some research has found interesting multilinguality in both public and proprietary LLMs \cite{qin2024multilingual} \cite{zhu2023multilingual}. Despite these developments, to the best of our knowledge, the capability of cross-lingual transfer in these LLMs has not been fully studied for sentiment analysis tasks, and it is still unclear how LLMs stand in comparison to existing multilingual pre-trained models in the cross-lingual transfer paradigm.

In this work, we examine a variety of pre-trained models and conduct a comprehensive study on the cross-lingual transfer capability in utterance-level sentiment analysis tasks with human speech transcript. We classify our candidate public pre-trained models into two categories: Small Multilingual Language Models (\textbf{SMLM})\footnote{We select SMLMs with fewer than 4B parameters in this work.}  such as XLM-R and mT5, and more recent Large Language Models (\textbf{LLM})\footnote{We select LLMs with at least 7B parameters in this work.}  primarily focused on English such as Llama-3 \cite{llama3modelcard} and Mistral \cite{jiang2023mistral}. In addition, we also include benchmarking with proprietary LLMs such as GPT-4 \cite{openai2024gpt4}, which is widely considered as the best LLM in terms of general capability. To avoid potential data contamination introduced in the pre-training process of recent LLMs \cite{sainz-etal-2023-nlp}, we curate and annotate proprietary sentiment datasets from in-house human conversation transcripts, and assess cross-lingual sentiment analysis from English to three target languages: Spanish, French and Chinese. Our evaluation results show that with the same supervised fine-tuning, SMLMs demonstrate superior zero-shot cross-lingual transfer capability even with much fewer model parameters. However, public LLMs exhibit rapid improvement in few-shot cross-lingual transfer scenarios and can surpass the performance of SMLMs when additional samples in the target language are provided. Our contributions of this research can be summarized in the following dimensions:

\begin{enumerate}
\item We provide a comprehensive comparison on fine-tuning-based cross-lingual transfer capability across a spectrum of public pre-trained language models, with up to 8 billion parameters in the sentiment analysis task on three human languages.
\item Our empirical findings show that some SMLMs (XLM-R, mT5) beat much larger public LLMs in zero-shot cross-lingual transfer. Nevertheless, larger LLMs surpass SMLMs and demonstrate stronger adaptation capability with few-shot fine-tuning in the target language. The best-performing SMLMs still show comparable performance to LLMs when more samples from the target language are provided.
\item We demonstrate that although proprietary GPT-3.5 and GPT-4 present the strongest performance in zero-shot cross-lingual sentiment analysis, with supervised fine-tuning, several public pre-trained language models can outperform GPT-3.5 and GPT-4 in sentiment analysis tasks with few-shot cross-lingual transfer.
\end{enumerate}

\begin{figure*}[t]
\centering
\includegraphics[width=1\textwidth, height=8cm]{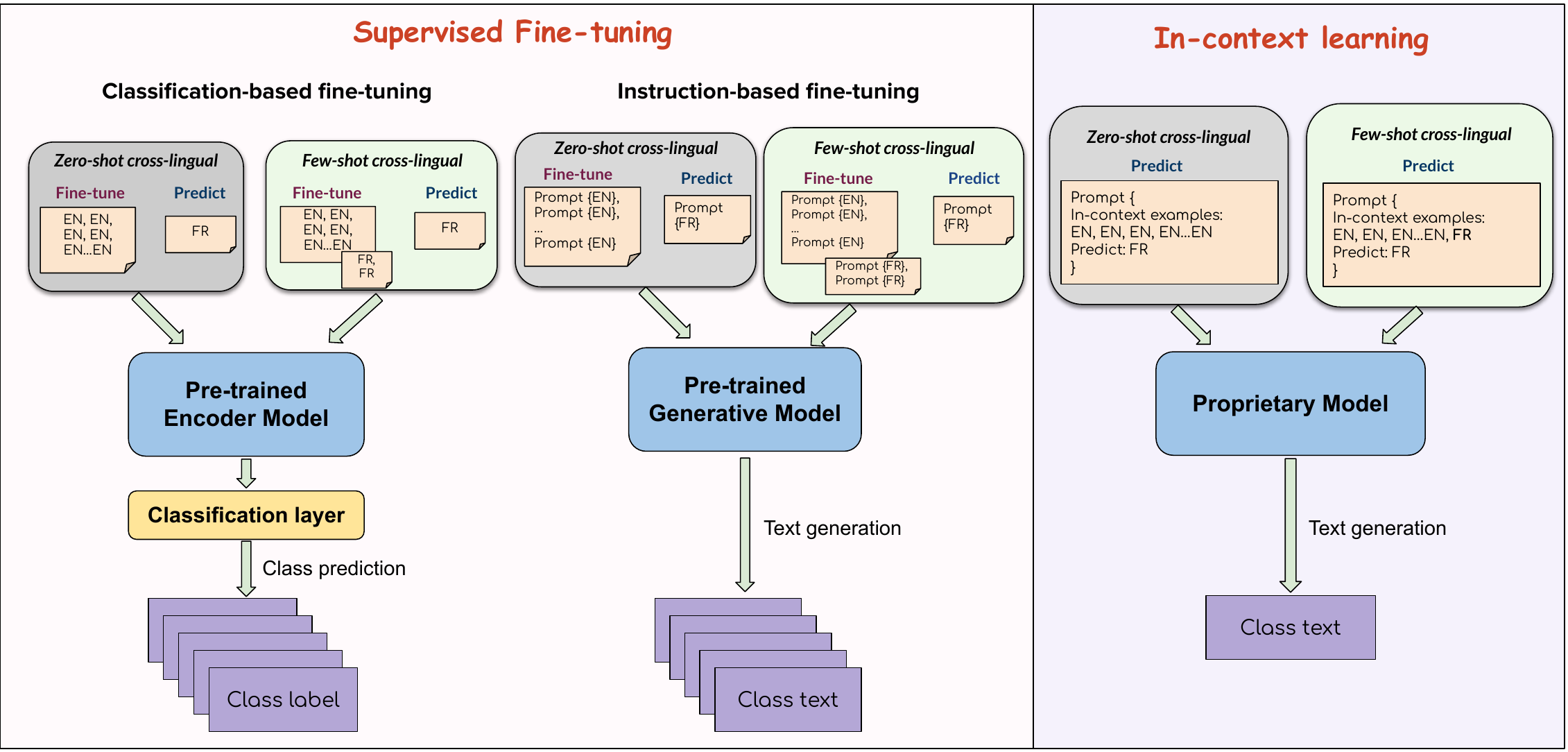}
\caption{\small Diagram of zero- and few- shot cross-lingual sentiment analysis from English (EN) to French (FR) under Supervised Fine-tuning (left) and In-context learning (right).}
\label{fig1}
\end{figure*}

\section{Background}
\subsection{Cross-lingual Sentiment Analysis}
\label{sec:2.1}
Sentiment analysis, as an important subfield of Natural Language Processing, concentrates on detecting and categorizing emotions and opinions in the text. Although the research predominantly focused on the English language initially, subsequent efforts have expanded to support cross-lingual sentiment analysis. This approach aims at leveraging one or several linguistically-rich source languages to enhance task performance in low-resource languages \cite{A-Survey-of-Cross-lingual}. Early methods such as \cite{Mining-Multilingual-Opinions} used Machine Translation for cross-lingual sentiment analysis, which became the mainstream methodology in the following years. Other studies focused on bridging the dataset disparities between source and target languages \cite{ZHANG2016129}, as well as generating parallel corpora for sentiment analysis tasks \cite{lu-etal-2011-joint} \cite{meng-etal-2012-cross}.  

The success of pre-trained models like BERT \cite{devlin-etal-2019-bert} has spurred adaptations for multilingual and cross-lingual applications, notably mBERT and XLM-R, which utilize a transformer encoder architecture and demonstrate strong capability in cross-lingual language understanding. These models are pre-trained with extensive multilingual corpora and subsequently fine-tuned for specific downstream tasks, thereby significantly enhancing sentiment analysis tasks across diverse languages \cite{barbieri-etal-2022-xlm}. \cite{xue-etal-2021-mt5} introduced mT5, which features a transformer encoder-decoder architecture and is pre-trained across over 101 languages, has shown superior performance in classification tasks such as XNLI \cite{conneau2018xnli} and surpassed both mBERT and XLM-R. More recently, advancements in unsupervised corpora and computational resources have facilitated the emergence of LLMs with a transformer decoder-only architecture, which have exhibited exceptional performance in various NLP tasks \cite{touvron2023llama} \cite{jiang2023mistral} \cite{brown2020language}. Despite these advancements, such LLMs are predominantly English-centric, and their multilingual capabilities remain somewhat ambiguous due to limited disclosure of training data specifics. Furthermore, the capabilities of cross-lingual transfer in these LLMs have yet to be thoroughly studied.

\subsection{Sentiment Analysis in Conversational Transcripts}
\label{sec:2.2}
Our work is situated within the context of human conversational transcript data; in our case, these transcript data are obtained from our internal company call centers, consisting of human-to-human conversations that mainly occur between a customer and a customer support agent. 

Analyzing such transcript data can be challenging to work with, even for English NLP models: conversational data contain mainly artifacts of spoken language, such as filler words, dysfluencies, and transcription errors by the automated speech recognition (ASR) model \cite{fu-etal-2022-entity}. Adding additional complexity by moving away from English-only data into other languages provides an opportunity to further test the limits of pre-trained language models: switching from one language to another does not always lend itself to a simple, one-to-one translation of each word -- especially in describing or expressing abstract concepts like sentiment. 

This complexity in cross-lingual sentiment analysis also comes from the need of considering both cultural and linguistic differences. For instance, one of our main observations on sentiment classification in real human conversation in Spanish was that Spanish speakers seem to focus on describing their complaint or situation instead of directly expressing their emotions. For example, they would rather say "{\fontfamily{cmr}\selectfont \textit{Esta es la quinta vez que los llamo}}" ("\textit{This is the fifth time I'm calling you guys}") instead of speaking up and expressing how frustrated they are with a simple and straightforward adjective, such as "{\fontfamily{cmr}\selectfont \textit{Estoy frustrado}}" ("\textit{I am frustrated}"). Whereas the statistical models will easily detect "{\fontfamily{cmr}\selectfont \textit{frustrado}}" and label it as negative sentiment, the abstract description that the speaker chooses in order to express their frustration in the first example will still present a challenge.

\begin{table*}[t]
\small
\centering
\begin{tabular}{p{1cm}p{3cm}p{3cm}p{3cm}p{3cm}}
&
English (EN) &
Spanish (ES) &
French (FR) &
Chinese (ZH) \\
\hline
\multirow{6}{*}{Neutral} &
We're busy, we can't complain, we're fine. &
Estamos ocupados, no podemos quejarnos, estamos bien. &
Nous sommes occupés, nous ne pouvons pas nous plaindre, nous allons bien. &
\begin{CJK*}{UTF8}{gbsn}
我们很忙，我们没什么要抱怨的，没事。
\end{CJK*}\\

&
There, I don't know why. &
Ahí, no sé por qué. &
Là, je ne sais pas pourquoi. &
\begin{CJK*}{UTF8}{gbsn}
这个，我不知道为什么。
\end{CJK*}\\
\hline

\multirow{7}{*}{Positive} &
I love the first one so I'm excited for this one, thanks. &
Me encanta el primero, así que estoy emocionado por este, gracias. &
J'adore le premier alors je suis excité pour celui-ci, merci. &
\begin{CJK*}{UTF8}{gbsn}
我很喜欢第一个，对此我感到很兴奋，谢谢。
\end{CJK*}\\

&
This is great, so professional, I'm sure the client was very impressed. &
Esto es genial, muy profesional, estoy seguro de que el cliente quedó muy impresionado. &
C'est génial, tellement professionnel, je suis sûr que le client était très impressionné. &
\begin{CJK*}{UTF8}{gbsn}
很好这非常专业，我相信客户一定印象非常深刻。
\end{CJK*}\\
\hline

\multirow{6}{*}{Negative} &
I think he's really pissed at me today. &
Creo que hoy está muy enojado conmigo. &
Je pense qu'il est vraiment très énervé contre moi aujourd'hui. &
\begin{CJK*}{UTF8}{gbsn}
我感觉他今天对我一定非常生气。
\end{CJK*}\\

&
Yes but I'm worried about being charged twice now. &
Sí, pero ahora me preocupa que me cobren dos veces. &
Oui mais je suis inquiet d'être facturé deux fois maintenant. &
\begin{CJK*}{UTF8}{gbsn}
是的，但我对于被收两次费用感到很担心。
\end{CJK*}\\

\end{tabular}

\caption{\small Examples of our proprietary sentiment datasets. }
\label{table1}
\end{table*}

\section{Methodology}
\label{sec:3}
\subsection{Supervised Fine-tuning}
\label{sec:3.1}
The objective of this work is to explore the cross-lingual transfer capability of pre-trained models within the context of a sentiment analysis task. To this end, we employ Supervised Fine-tuning (\textbf{SFT}) on publicly available pre-trained models using annotated proprietary sentiment datasets (detailed in Section \ref{sec:4.1}). Each model is fine-tuned to categorize sentiments as Positive, Negative, or Neutral based on the input provided. Given the diversity in pre-training objectives among different models, we implement two distinct fine-tuning approaches illustrated in Figure \ref{fig1}, which are tailored to the architecture of the pre-trained models:

\begin{itemize}
\item \textbf{Classification-based fine-tuning}: applicable to transformer encoder-only models such as mBERT and XLM-R, we add a classification layer on top of the pre-trained models and fine-tune the model to directly predict a sentiment class.
\item \textbf{Instruction-based fine-tuning}: used for transformer encoder-decoder (e.g. mT5) and decoder-only (e.g. Llama-3) structures, we construct an instruction to prompt the model to generate a text output corresponding to a sentiment class. The specific prompt format is detailed in Appendix \ref{sec:fine-tuning-prompt}.
\end{itemize}

To comprehensively evaluate the cross-lingual transfer capabilities of these pre-trained models through fine-tuning, we target both zero- and few- shot cross-lingual transfer from a source to a target language. In \textit{Zero-shot Cross-lingual Transfer} setting, the model is fine-tuned exclusively with an annotated dataset in the source language and subsequently tasked with making predictions in a target language. Note that for generative tasks, merely input language alteration is applied while the instruction component remains constant. \textit{Few-shot Cross-lingual Transfer} extends the zero-shot framework by additionally incorporating \textbf{\textit{N}} labeled examples from the target language into the fine-tuning process, alongside the source language dataset. The format of the prompt used remains consistent with zero-shot for generative tasks, detailed in Appendix \ref{sec:fine-tuning-prompt}.

\begin{table*}[t]
\small
\centering
\begin{tabular}{ccccc}

Model type &
Name &
Architecture &
\# of param. &
Claimed language support
\\
\hline
\multirow{ 6}{*}{SMLM} &
mBERT &
encoder &
110M &
104 langs\\
&
XLM-R-base &
encoder &
250M &
100 langs\\
&
XLM-R-large &
encoder &
560M &
100 langs\\
&
mT5-base &
encoder-decoder &
580M &
101 langs\\
&
mT5-large &
encoder-decoder &
1.2B &
101 langs\\
&
mT5-xl &
encoder-decoder &
3.7B &
101 langs\\
\hline
\multirow{ 4}{*}{English-centric LLM} &
Mistral-7B &
decoder & 
7B &
Unclear \\
&
Falcon-7B &
decoder & 
7B &
Mainly EN, DE, ES, FR \\
&
Llama2-7B &
decoder & 
7B &
Intended for EN \\
&
Llama3-8B &
decoder & 
8B &
Intended for EN \\

\end{tabular}

\caption{\small List of public pre-trained models evaluated in our experiments. }
\label{table2}
\end{table*}

\subsection{In-context Learning}
\label{sec:3.2}
Recent advancements have highlighted in-context learning as a viable alternative to the traditional fine-tuning approach for generative models \cite{dong2023survey}. Due to the access limitation and our data privacy policy, we are not able to fine-tune proprietary LLMs using our proprietary datasets. Consequently, we employ in-context learning through the prompt to simulate an experiment setting as conducting SFT on public models. Nonetheless, the inherent limitation regarding the context length in various close source LLMs poses a challenge; these models may not accommodate as many examples within a prompt as is feasible for SFT in open source counterparts. Figure \ref{fig1} shows an illustrative diagram of in-context learning for this sentiment analysis task. 

To assess cross-lingual transfer capabilities as Section \ref{sec:3.1} through in-context learning, we construct in-context examples with different sources of languages accordingly. Specifically, for \textit{Zero-shot Cross-lingual Transfer}, the prompts include examples solely from the source language. In contrast, for \textit{Few-shot Cross-lingual Transfer}, additional supplementary examples in the target language are also applied. Prompts with in-context examples we use to evaluate proprietary LLMs are attached in Appendix \ref{sec:in-context-prompt}.

\section{Experiment}
\label{sec:4}
In this section, we first present a detailed description of our internal proprietary sentiment datasets which are used for fine-tuning and evaluation. Then, we provide necessary introductions to a diverse array of public pre-trained models we will study for this work. Finally, we show the hardware and software resources employed in conducting the experiment.

\subsection{Dataset}
\label{sec:4.1}
The proprietary datasets used in this study are utterance-level sentiment data for four languages: English, Spanish, French, Chinese (Table \ref{table1}). Utterance boundaries are generated by our in-house ASR system when a short pause or speaker change is detected in the audio stream. We randomly sampled English and Spanish utterances from the real conversational transcript from our call center applications and each instance is labeled as \textbf{Positive}, \textbf{Negative} or \textbf{Neutral} by human annotators. The annotation was done via a third-party vendor, allowing us to configure our ontology and direct the annotators to select the appropriate category for the sentiment detected in each utterance according to guidelines we developed. Our guidelines include definitions for each sentiment as well as a broad list of examples (a gold dataset manually annotated by our internal team). Inter-annotator agreement is calculated automatically by our annotation vendor, and a high agreement threshold is applied to ensure the quality of the annotation results.\footnote{\url{https://docs.labelbox.com/docs/consensus}}

Constrained by resources, we are not able to sample and annotate French and Chinese datasets under the same setting. Instead, we leverage machine translation (through GPT-4, detailed in Appendix \ref{sec:machine-translation}) to create parallel French and Chinese datasets based on the annotated English counterpart. All machine-translated datasets were reviewed by speakers of the target language to ensure that the translations were comparable to the original English. There were some minor issues identified in the machine-translated data during review: namely, occasionally GPT-4 refuses to translate a sample, producing a refusal in the target language instead, or it produced a commentary on the English transcript in the target language in lieu of translating it directly. These samples were identified and removed, and the remaining samples were deemed to be accurate translations by the speakers of the target languages.

As our objective is to study the cross-lingual sentiment analysis from English to target languages, we assemble English data with a much larger size, while Spanish, French and Chinese with a limited amount sufficient only to support few-shot learning and testing purposes. A summary of the total amount of data used for the following experiment is as follows:

\begin{itemize}
\item[-] English: 30,000 instances for fine-tuning, 3,000 for development.
\item[-] Spanish: 600 instances for fine-tuning and 3,000 for testing.
\item[-] French: 600 instances for fine-tuning and 3,000 for testing.
\item[-] Chinese: 600 instances for fine-tuning and 3,000 for testing.
\end{itemize}
where we ensure sentiment labels are uniformly distributed across all sets.

Table \ref{table1} shows exemplary cases of our proprietary datasets in different languages, providing insight into domain-specific textual characteristics. It is worth mentioning that these examples have no identifying information and are intended for illustrative purposes only. The use of internal call transcript data ensures that all model evaluations are immune from unintended data contamination of the pre-trained models, which could otherwise lead to an overestimation of their performance \cite{sainz-etal-2023-nlp}.

\begin{table*}[t]
\small
\centering
\renewcommand{\arraystretch}{1.3}
\begin{tabularx}{1\textwidth} { 
  | >{\centering\arraybackslash}X
  | >{\centering\arraybackslash}X 
   >{\centering\arraybackslash}X
   >{\centering\arraybackslash}X
   >{\centering\arraybackslash}X
   >{\centering\arraybackslash}X
   >{\centering\arraybackslash}X
  | >{\centering\arraybackslash}X 
   >{\centering\arraybackslash}X
   >{\centering\arraybackslash}X
   >{\centering\arraybackslash}X
  | >{\centering\arraybackslash}X 
   >{\centering\arraybackslash}X| }
\hline
&
\multicolumn{6}{c|}{Public SMLM} & 
\multicolumn{4}{c|}{Public LLM} &
\multicolumn{2}{c|}{Proprietary LLM}\\
&
\multicolumn{6}{c|}{\scriptsize Supervised Fine-tuning} & 
\multicolumn{4}{c|}{\scriptsize Supervised Fine-tuning} &
\multicolumn{2}{c|}{\scriptsize In-context Learning}\\
&
\textbf{\scriptsize{mBERT}} &
\textbf{\scriptsize{XLM-R-base}} &
\textbf{\scriptsize{XLM-R-large}}  &
\textbf{\scriptsize{mT5-base}}  &
\textbf{\scriptsize{mT5-large}}  &
\textbf{\scriptsize{mT5-xl}} &
\textbf{\scriptsize{Mistral}} &
\textbf{\scriptsize{Falcon}}  &
\textbf{\scriptsize{Llama-2}}  &
\textbf{\scriptsize{Llama-3}} &
\textbf{\scriptsize{GPT-3.5}} &
\textbf{\scriptsize{GPT-4}}
\\
&
\scriptsize{110M} &
\scriptsize{250M} &
\scriptsize{560M}  &
\scriptsize{580M}  &
\scriptsize{1.2B}  &
\scriptsize{3.7B} &
\scriptsize{7B} &
\scriptsize{7B}  &
\scriptsize{7B}  &
\scriptsize{8B} &
\scriptsize{-} &
\scriptsize{-}
\\
\hline
ES &
47.1  &
54.4  &
58.7  &
60.2  &
63.4  &
60.0  &
44.8  &
55.3  &
60.1  &
57.9  &
75.6  &
74.8 \\
FR &
45.3  &
71.8  &
76.8  &
75.4  &
79.7  &
73.8  &
48.4  &
70.7  &
74.5  &
77.4  &
80.3  &
79.3 \\
ZH &
54.2  &
72.3  &
76.9  &
74.8  &
77.3  &
71.5  &
40.4  &
71.9  &
64.9  &
73.3  &
82.3  &
80.2 \\
Avg &
48.9  &
66.2  &
70.8  &
70.1  &
\textbf{73.5} &
68.4  &
44.5  &
66.0  &
66.5  &
69.5  &
\textbf{79.4}  &
\textbf{78.1} \\
\hline
\end{tabularx}
\caption{\small F1 score comparison in zero-shot cross-lingual transfer on our proprietary sentiment analysis datasets. ES: Spanish, FR: French, ZH: Chinese. Top-3 average F1 scores are marked in bold.}
\label{table3}
\end{table*}

\begin{figure*}[t]
\centering
\includegraphics[width=1\textwidth, height=6.5cm]{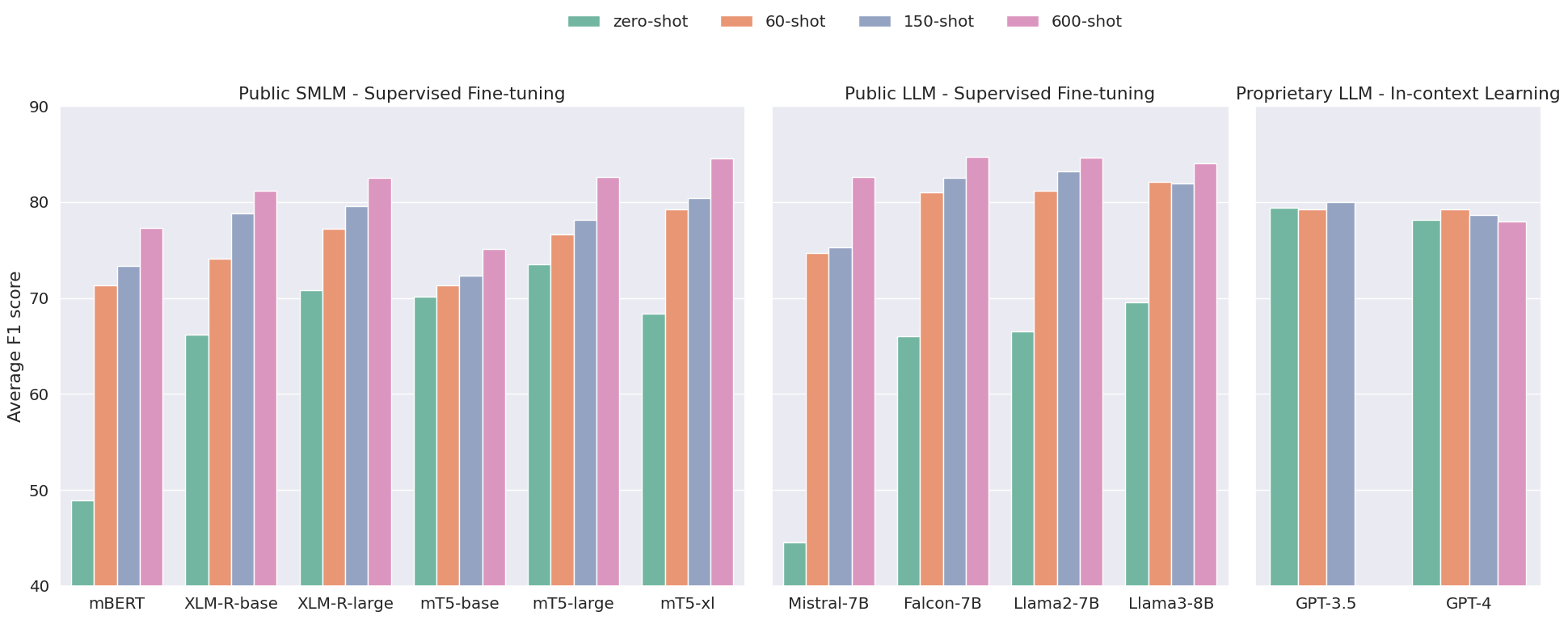}
\caption{\small Average F1 score performance comparison (across ES, FR and ZH) under N-shot settings. GPT-3.5 is not included in this 600-shot due to the context length limit.}
\label{fig2}
\end{figure*}

\subsection{Selected pre-trained Models}
\label{sec:4.2}
In this work, we investigate a variety of public pre-trained language models, with a range of sizes and architectures. For SMLM, we have selected models from mBERT, XLM-R and mT5 model families with up to 3.7 billion parameters. All models in our SMLM selection are known for their support for over 100 human languages and have demonstrated efficacy in tasks that require multilingual and cross-lingual capabilities, as evidenced by references \cite{doddapaneni2021primer} \cite{xue-etal-2021-mt5}. For English-centric LLMs, the details are little disclosed regarding the specific human languages incorporated during the pre-training phase. Therefore, we include the most prominent and widely recognized models from Llama family and Mistral with 7 to 8 billion parameters sizes. In additional, Falcon-7B is also added to our analysis as it explicitly claims proficiency in German, Spanish and French in addition to English. The specifics of all the pre-trained models utilized in our experiments are detailed in Table \ref{table2}.

\subsection{Experiment Setup}
\label{sec:4.3}
The fine-tuning and inference processes for our model are conducted using the Huggingface framework \cite{wolf2020huggingfaces} on a single-node Linux system equipped with eight Nvidia A100 80G GPUs. For experiments on proprietary LLMs, we use “{\fontfamily{qcr}\selectfont{\normalsize{gpt-3.5-turbo-0125}}}” endpoint for GPT-3.5 and “{\fontfamily{qcr}\selectfont{\normalsize{gpt-4-1106-preview}}}” endpoint for GPT-4.

In order to ensure deterministic output from generative models, temperature is set as 0 for all public and proprietary models in our experiments.

\section{Results}
\label{sec:5}
To facilitate a comprehensive comparison between SMLMs and LLMs on cross-lingual sentiment analysis, we follow the zero-shot and few-shot cross-lingual fine-tuning methodologies described in \ref{sec:3.1} and evaluate the model performance respectively. The F1 score (micro) is employed as the accuracy evaluation metric in the following sentiment analysis experiments.

\subsection{Zero-shot Cross-lingual Transfer}
\label{sec:5.1}
We first fine-tune public pre-trained models in zero-shot cross-lingual transfer setting through SFT as detailed in Section \ref{sec:3.1}, exposed to only the English fine-tuning dataset described in \ref{sec:4.1}. Note that we leverage in-context learning for proprietary LLMs as discussed in Section \ref{sec:3.2}. However, due to constraints on context length, these proprietary LLMs are not exposed to the entirety of the English fine-tuning set; instead, they are prompted with a set of 300 examples, carefully balanced across different classes for this experiment.

Evaluation results are presented in Table \ref{table3}. It is clear that both GPT-3.5 and GPT-4 exhibit significant advantages over fine-tuned public models on target languages in zero-shot. Surprisingly, among the public models, several SMLMs such as XLM-R-large (560M), mT5-base (580M) and mT5-large (1.2B), show better zero-shot cross-lingual transfer capability compared to the considerably larger Mistral-7B, Falcon-7B, Llama2-7B and Llama3-7B models.  In particular, mT5-large surpasses all other open source candidates by a substantial margin across all testing languages despite having only 1.2 billion parameters.

\subsection{Few-shot Cross-lingual Transfer}
\label{sec:5.2}
We then fine-tune and evaluate public models under the few-shot cross-lingual transfer setting described in Section \ref{sec:3}, where we randomly select \textbf{\textit{N}} training samples in the target language and use them in fine-tuning in conjunction with the English fine-tuning data. In order to better investigate the adaptability of the models, we vary \textbf{\textit{N}} among \{60, 150, 600\}, thereby conducting \textbf{60-shot}, \textbf{150-shot} and \textbf{600-shot} experiments respectively. The selection of these three values provides a wide spectrum for comparative analysis, also ensures a sufficient representation while maintaining resource-efficient. For proprietary LLMs, an additional \textbf{\textit{N}} samples in target language are appended to the prompt  during in-context learning to establish a similar few-shot cross-lingual setup. 

The evaluation results of average F1 scores across three target languages (ES, FR and ZH) are presented in Figure \ref{fig2}, under the settings of 60-shot, 150-shot and 600-shot. Detailed F1 scores per language are also provided in Appendix \ref{sec:per-lang-eval}. Our observations and findings can be summarized as follows:

\begin{enumerate}[i]
    \item Among public pre-trained models, despite their underperformance relative to SMLMs in zero-shot cross-lingual transfer as evidenced in Table \ref{table3}, English-centric LLMs present strong adaptation capability in few-shot cross-lingual sentiment analysis. Notably, all public LLMs exhibit significant relative improvements compared to their zero-shot performance. It is worth pointing out that with 60-shot and 150-shot, LLMs such as Falcon-7B, Llama2-7B and Llama3-8B surpass the performance of all SMLMs by a considerable margin. The only exception is Mistral-7B, which is still outperformed by several SMLMs with few-shot.
    \item With an increased volume of training data in the target language, specifically under 600-shot condition, mT5-xl with 3.7B parameters has a comparable performance to the much larger  Falcon-7B,  Llama2-7B and Llama3-8B models.
    \item Contrary to their dominance in the zero-shot cross-lingual setting, GPT-4 and GPT-3.5 exhibit very limited improvement in few-shot cross-lingual sentiment analysis with in-context examples. Several public models are capable of surpassing these prominent proprietary LLMs following fine-tuning.

\end{enumerate}

\section{Conclusion}
In this study, we explore the capabilities of cross-lingual sentiment analysis across a variety of pre-trained language models.  We show that smaller XLM-R-large (560M), mT5-base (580M) and mT5-large (1.2B) have superior zero-shot cross-lingual transfer capabilities compared to the considerably larger Mistral-7B, Falcon-7B, Llama2-7B and Llama3-8B models. This highlights the efficiency and potential of Small Multilingual Language Models (SMLM) for sentiment analysis in low-resource languages. On the other hand, our findings reveal that the larger English-centric LLMs like Falcon-7B and Llama2-7B can quickly adapt and show much improved performance with a few-shot cross-lingual setup, which indicates their robustness in learning from limited data from the target language. Moreover, proprietary LLMs such as GPT-3.5 and GPT-4 exhibit the strongest zero-shot performance in cross-lingual sentiment analysis tasks, however, in scenarios involving few-shot learning, several fine-tuned public pre-trained models are able to surpass these proprietary giants. 

\section{Limitation}
Although our findings in this study appear to be consistent in all target languages tested, due to the limitation of our resources, it is still unclear how the models would behave in other low-resource languages with even less appearance during pre-training. In addition, due to the incomparable model sizes, we are not able to draw any conclusions on whether model architecture difference (transformer encoder-only, decoder-only and encoder-decoder) could play a role in cross-lingual sentiment analysis capabilities. Further research could be extended in these directions.

\bibliography{custom}

\appendix

\section{Appendix}
\label{sec:appendix}

\subsection{Prompt Format for Supervised Fine-tuning}
\label{sec:fine-tuning-prompt}
We employ the following prompt format in supervised fine-tuning for public generative models:
\\

{\fontfamily{qcr}\selectfont{\normalsize \small Below is an utterance extracted from the transcript of a business call, identify the speaker's sentiment in this utterance.
The sentiment should be one of the following: \\
"Positive": The speaker expresses favorable emotions and mental states, for example, euphoria and joy, happiness, excitement, fascination, satisfaction, pride, gratitude, relief, surprise, etc. \\
"Negative": The speaker expresses unfavorable emotions and mental states, for example, disgust, sadness, disappointment, worry, insecurity, annoyance, fury, anger, fear, depression, frustration, etc.\\
"Neutral": Statement in which the speaker does not express emotions, but in which a fact is simply stated and no explicit emotions or feelings are conveyed.\\
    What is the sentiment in the following utterance? Only respond with the sentiment without explanation: \\
    \#\#\# Input: \{\textbf{utterance text}\} \\
    \#\#\# Output:
    }}\\

\subsection{Prompt Format for In-context Learning}
\label{sec:in-context-prompt}
The following prompt with in-context examples is used for calling proprietary LLM APIs:\\

{\fontfamily{qcr}\selectfont{\normalsize \small Below is an utterance extracted from the transcript of a business call, identify the speaker's sentiment in this utterance.
The sentiment should be one of the following: \\
"Positive": The speaker expresses favorable emotions and mental states, for example, euphoria and joy, happiness, excitement, fascination, satisfaction, pride, gratitude, relief, surprise, etc. \\
"Negative": The speaker expresses unfavorable emotions and mental states, for example, disgust, sadness, disappointment, worry, insecurity, annoyance, fury, anger, fear, depression, frustration, etc.\\
"Neutral": Statement in which the speaker does not express emotions, but in which a fact is simply stated and no explicit emotions or feelings are conveyed.\\
    Here are some examples:\\
    \#\#\# Input: \{utterance text 1\}\\
    \#\#\# Output: \{sentiment label 1\} \\
    \\
    \#\#\# Input: \{utterance text 2\}\\
    \#\#\# Output: \{sentiment label 2\}\\
    \\
    \#\#\# Input: \{utterance text 3\}\\
    \#\#\#   Output: \{sentiment label 3\}\\
    \\
    ...\\
    \\
    What is the sentiment in the following utterance? Only respond with the sentiment without explanation: \\
    \#\#\# Input: \{\textbf{utterance text}\} \\
    \#\#\# Output:
    }}\\

\subsection{Machine translation details}
\label{sec:machine-translation}

The machine translation process described in Section \ref{sec:4.1} utilizes GPT-4 endpoint “{\fontfamily{qcr}\selectfont{\normalsize{gpt-4-1106-preview}}}”. The prompt used for machine translation is as follows:\\
{\fontfamily{qcr}\selectfont{\normalsize \small Below is a transcribed utterance from human conversations, translate it from English to \{TARGET\_LANG\}: \\
    \#\#\# Input: \{\textbf{English utterance}\} \\
    \#\#\# Output:
    }}\\

TARGET\_LANG refers to the target languages in our machine translation process, i.e. French and Chinese.

\subsection{Per-language Evaluation Tables for Few-shot Cross-lingual}
\label{sec:per-lang-eval}

Supplementary to Section \ref{sec:5.2}, detailed per language evaluation results on few-shot cross-lingual are listed in Table \ref{table4}, Table \ref{table5}, and Table \ref{table6}

\begin{table*}[t]
\small
\centering
\renewcommand{\arraystretch}{1.3}
\begin{tabularx}{1\textwidth} { 
  | >{\centering\arraybackslash}X
  | >{\centering\arraybackslash}X 
   >{\centering\arraybackslash}X
   >{\centering\arraybackslash}X
   >{\centering\arraybackslash}X
   >{\centering\arraybackslash}X
   >{\centering\arraybackslash}X
  | >{\centering\arraybackslash}X 
   >{\centering\arraybackslash}X
   >{\centering\arraybackslash}X
   >{\centering\arraybackslash}X
  | >{\centering\arraybackslash}X 
   >{\centering\arraybackslash}X| }
\hline
&
\multicolumn{6}{c|}{Public SMLM} & 
\multicolumn{4}{c|}{Public LLM} &
\multicolumn{2}{c|}{Proprietary LLM}\\
&
\multicolumn{6}{c|}{\scriptsize Supervised Fine-tuning} & 
\multicolumn{4}{c|}{\scriptsize Supervised Fine-tuning} &
\multicolumn{2}{c|}{\scriptsize In-context Learning}\\
&
\textbf{\scriptsize{mBERT}} &
\textbf{\scriptsize{XLM-R-base}} &
\textbf{\scriptsize{XLM-R-large}}  &
\textbf{\scriptsize{mT5-base}}  &
\textbf{\scriptsize{mT5-large}}  &
\textbf{\scriptsize{mT5-xl}} &
\textbf{\scriptsize{Mistral}} &
\textbf{\scriptsize{Falcon}}  &
\textbf{\scriptsize{Llama-2}}  &
\textbf{\scriptsize{Llama-3}} &
\textbf{\scriptsize{GPT-3.5}} &
\textbf{\scriptsize{GPT-4}}
\\
&
\scriptsize{110M} &
\scriptsize{250M} &
\scriptsize{560M}  &
\scriptsize{580M}  &
\scriptsize{1.2B}  &
\scriptsize{3.7B} &
\scriptsize{7B} &
\scriptsize{7B}  &
\scriptsize{7B}  &
\scriptsize{8B} &
\scriptsize{-} &
\scriptsize{-}
\\
\hline
ES &
71.0  &
62.7  &
67.1  &
59.7  &
65.3  &
73.2  &
73.1  &
76.8  &
77.7  &
77.6  &
76.0  &
76.8 \\
FR &
69.3  &
79.7  &
82.7  &
76.1  &
83.7  &
83.8  &
76.1  &
82.3  &
84.7  &
85.2  &
81.6  &
80.3 \\
ZH &
73.7  &
80.0  &
81.7  &
78.0  &
80.8  &
80.7  &
74.9  &
84.0  &
81.2  &
83.5  &
80.1  &
80.4 \\
Avg &
71.3  &
74.1  &
77.2  &
71.3  &
76.6 &
79.2  &
74.7  &
\textbf{81.0}  &
\textbf{81.2}  &
\textbf{82.1}  &
79.2  &
79.2 \\
\hline
\end{tabularx}
\caption{\small F1 score comparison in \textbf{60-shot} cross-lingual transfer on our proprietary sentiment analysis datasets. ES: Spanish, FR: French, ZH: Chinese. Top-3 average F1 scores are marked in bold.}
\label{table4}
\end{table*}

\begin{table*}[t]
\small
\centering
\renewcommand{\arraystretch}{1.3}
\begin{tabularx}{1\textwidth} { 
  | >{\centering\arraybackslash}X
  | >{\centering\arraybackslash}X 
   >{\centering\arraybackslash}X
   >{\centering\arraybackslash}X
   >{\centering\arraybackslash}X
   >{\centering\arraybackslash}X
   >{\centering\arraybackslash}X
  | >{\centering\arraybackslash}X 
   >{\centering\arraybackslash}X
   >{\centering\arraybackslash}X
   >{\centering\arraybackslash}X
  | >{\centering\arraybackslash}X 
   >{\centering\arraybackslash}X| }
\hline
&
\multicolumn{6}{c|}{Public SMLM} & 
\multicolumn{4}{c|}{Public LLM} &
\multicolumn{2}{c|}{Proprietary LLM}\\
&
\multicolumn{6}{c|}{\scriptsize Supervised Fine-tuning} & 
\multicolumn{4}{c|}{\scriptsize Supervised Fine-tuning} &
\multicolumn{2}{c|}{\scriptsize In-context Learning}\\
&
\textbf{\scriptsize{mBERT}} &
\textbf{\scriptsize{XLM-R-base}} &
\textbf{\scriptsize{XLM-R-large}}  &
\textbf{\scriptsize{mT5-base}}  &
\textbf{\scriptsize{mT5-large}}  &
\textbf{\scriptsize{mT5-xl}} &
\textbf{\scriptsize{Mistral}} &
\textbf{\scriptsize{Falcon}}  &
\textbf{\scriptsize{Llama-2}}  &
\textbf{\scriptsize{Llama-3}} &
\textbf{\scriptsize{GPT-3.5}} &
\textbf{\scriptsize{GPT-4}}
\\
&
\scriptsize{110M} &
\scriptsize{250M} &
\scriptsize{560M}  &
\scriptsize{580M}  &
\scriptsize{1.2B}  &
\scriptsize{3.7B} &
\scriptsize{7B} &
\scriptsize{7B}  &
\scriptsize{7B}  &
\scriptsize{8B} &
\scriptsize{-} &
\scriptsize{-}
\\
\hline
ES &
71.9  &
71.6  &
71.8  &
60.5  &
69.4  &
74.7  &
71.1  &
76.8  &
79.7  &
77.6  &
76.3  &
74.5 \\
FR &
71.3  &
82.0  &
82.9  &
78.0  &
83.3  &
83.0  &
76.0  &
86.1  &
84.2  &
82.9  &
81.9  &
78.7 \\
ZH &
76.8  &
82.7  &
84.1  &
78.4  &
81.7  &
83.6  &
78.7  &
84.5  &
85.6  &
85.2  &
81.7  &
82.6 \\
Avg &
73.3  &
78.8  &
79.6  &
72.3  &
78.1 &
80.4  &
75.3  &
\textbf{82.5}  &
\textbf{83.2}  &
\textbf{81.9}  &
80.0  &
78.6 \\
\hline
\end{tabularx}
\caption{\small F1 score comparison in \textbf{150-shot} cross-lingual transfer on our proprietary sentiment analysis datasets. ES: Spanish, FR: French, ZH: Chinese. Top-3 average F1 scores are marked in bold.}
\label{table5}
\end{table*}

\begin{table*}[t]
\small
\centering
\renewcommand{\arraystretch}{1.3}
\begin{tabularx}{1\textwidth} { 
  | >{\centering\arraybackslash}X
  | >{\centering\arraybackslash}X 
   >{\centering\arraybackslash}X
   >{\centering\arraybackslash}X
   >{\centering\arraybackslash}X
   >{\centering\arraybackslash}X
   >{\centering\arraybackslash}X
  | >{\centering\arraybackslash}X 
   >{\centering\arraybackslash}X
   >{\centering\arraybackslash}X
   >{\centering\arraybackslash}X
  | >{\centering\arraybackslash}X
  >{\centering\arraybackslash}X| }
\hline
&
\multicolumn{6}{c|}{Public SMLM} & 
\multicolumn{4}{c|}{Public LLM} &
\multicolumn{2}{c|}{Proprietary LLM}\\
&
\multicolumn{6}{c|}{\scriptsize Supervised Fine-tuning} & 
\multicolumn{4}{c|}{\scriptsize Supervised Fine-tuning} &
\multicolumn{2}{c|}{\scriptsize In-context Learning}\\
&
\textbf{\scriptsize{mBERT}} &
\textbf{\scriptsize{XLM-R-base}} &
\textbf{\scriptsize{XLM-R-large}}  &
\textbf{\scriptsize{mT5-base}}  &
\textbf{\scriptsize{mT5-large}}  &
\textbf{\scriptsize{mT5-xl}} &
\textbf{\scriptsize{Mistral}} &
\textbf{\scriptsize{Falcon}}  &
\textbf{\scriptsize{Llama-2}}  &
\textbf{\scriptsize{Llama-3}} &
\textbf{\scriptsize{GPT-3.5}} &
\textbf{\scriptsize{GPT-4}}
\\
&
\scriptsize{110M} &
\scriptsize{250M} &
\scriptsize{560M}  &
\scriptsize{580M}  &
\scriptsize{1.2B}  &
\scriptsize{3.7B} &
\scriptsize{7B} &
\scriptsize{7B}  &
\scriptsize{7B}  &
\scriptsize{8B} &
\scriptsize{-} &
\scriptsize{-}
\\
\hline
ES &
74.0  &
74.0  &
77.4  &
64.4  &
77.9  &
77.6  &
76.3  &
79.0  &
79.0  &
76.6  &
- &
73.9  \\
FR &
76.1  &
83.7  &
83.8  &
79.9  &
86.2  &
87.4  &
83.6  &
86.6  &
86.8  &
86.1  &
- &
78.8 \\
ZH &
81.8  &
85.8  &
86.4  &
80.9  &
83.8  &
88.6  &
87.8  &
88.3  &
88.0  &
89.3  &
- &
81.4 \\
Avg &
77.3  &
81.2  &
82.5  &
75.1  &
82.6 &
\textbf{84.5}  &
82.6  &
\textbf{84.7}  &
\textbf{84.6}  &
84.0  &
- &
78.0 \\
\hline
\end{tabularx}
\caption{\small F1 score comparison in \textbf{600-shot} cross-lingual transfer on our proprietary sentiment analysis datasets. ES: Spanish, FR: French, ZH: Chinese. Top-3 average F1 scores are marked in bold. GPT-3.5 is not included in this evaluation due to the context length limit.}
\label{table6}
\end{table*}

\end{document}